\pdfoutput=1
\documentclass[11pt]{article}

\usepackage[]{EMNLP2023}

\usepackage{graphicx}
\usepackage{geometry}
\usepackage{listings} % code
\usepackage{url}
\usepackage{microtype}
\usepackage{amsmath,amssymb}
\usepackage{times}
\usepackage{latexsym}
\usepackage{graphicx}
\usepackage{multirow}
\usepackage{booktabs}
\usepackage{tablefootnote}
\usepackage{colortbl} %\cellcolor
\usepackage{paralist} %\begin{compactitem}
\usepackage{soul} %\sethlcolor
\usepackage{subfigure}
\usepackage{xspace}
\usepackage{stfloats}
\usepackage{microtype}

\definecolor{codegreen}{rgb}{0,0.6,0}
\definecolor{codegray}{rgb}{0.5,0.5,0.5}
\definecolor{codepurple}{rgb}{0.58,0,0.82}
\definecolor{backcolour}{RGB}{251, 253, 253}
\definecolor{codeorange}{RGB}{206, 127, 71}

\lstdefinestyle{mystyle}{
    backgroundcolor=\color{backcolour},
    commentstyle=\color{codegreen},
    keywordstyle=\color{codepurple},
    numberstyle=\tiny\color{codegray},
    stringstyle=\color{codeorange},
    basicstyle=\ttfamily\footnotesize,
    breakatwhitespace=false,
    breaklines=true,
    captionpos=b,
    keepspaces=true,
    numbers=left,
    numbersep=5pt,
    showspaces=false,
    showstringspaces=false,
    showtabs=false,
    tabsize=2
}

\usepackage[T1]{fontenc}
\usepackage[utf8]{inputenc}

\usepackage{microtype}

\usepackage{inconsolata}

\usepackage{hyperref}
\usepackage{xcolor}
\usepackage{pifont}
\usepackage{soul}
\usepackage{url}
\usepackage[T1]{fontenc}
\usepackage{siunitx}
\usepackage{graphicx}
\usepackage{float}
\usepackage{multirow}
\usepackage{color}
\usepackage{wrapfig}
\usepackage{booktabs}
\usepackage{amssymb}
\usepackage{subfigure}
\usepackage{listings}
\usepackage{makecell}
\usepackage{amssymb,mathrsfs,amsmath}
\usepackage{amsmath, bm}
\usepackage{colortbl}
\definecolor{xred}{HTML}{BD4242}
\definecolor{xblue}{HTML}{C7A085}
\definecolor{xblues}{HTML}{52B256}
\definecolor{xgreen}{HTML}{52B256}
\definecolor{xpurple}{HTML}{7F52B2}
\definecolor{xorange}{HTML}{FD9337}
\definecolor{xdotted}{HTML}{999999}
\definecolor{xgray}{HTML}{777777}
\definecolor{xcyan}{HTML}{80F5DC}
\definecolor{xpink}{HTML}{f690ea}
\definecolor{xgraycyan}{HTML}{82bceb}
\usepackage{tikz}
\usetikzlibrary{positioning, calc}
\usepackage[most]{tcolorbox}

\tcbset{
	common/.style n args={2}{
		colframe={#1},
		colback={#1!5},
		colbacktitle={#1},
		lower separated=false,
		coltitle=white,
		boxrule=1pt,
		fonttitle=\bfseries,
		enhanced,
		breakable,
		top=8pt,
		before skip=8pt,
		attach boxed title to top left={
			yshift=-0.25cm,
			xshift=0.38cm,
		},
		boxed title style={
			boxrule=0pt,
			colframe=white,
			arc=5pt,
			outer arc=4pt,
		},
		separator sign={~~},
		overlay unbroken and last={
			\node[text=white, align=right, rounded corners, fill=#1, xshift=-4mm, minimum height=6mm, anchor=east] at (frame.south east) {#2};
		}
	},
	defstyle/.style={
		common={xpurple}{$\bm{\pi}$},
	},
	theoremstyle/.style={
		common={xgraycyan}{$\star$},
	},
	proofstyle/.style={
		common={xgreen}{\textbf{QED}},
	},
	examplestyle/.style={
		common={xblue}{$\bigstar$},
	},
    examplestyles/.style={
		common={xblues}{$\bigstar$},
	},
	notestyle/.style={
		common={xred}{\textbf{!}},
	},
	challangestyle/.style={
		common={xorange}{\textbf{?}},
	},
}

\newtcbtheorem{definition}{Definition}{defstyle}{def}
\newtcbtheorem{theorem}{Theorem}{theoremstyle}{theorem}
\newtcbtheorem{proof}{Proof}{proofstyle}{proof}
\newtcbtheorem{example}{Example}{examplestyle}{example}
\newtcbtheorem{examples}{Examples}{examplestyle}{example}
\newtcbtheorem{note}{Note}{notestyle}{note}
\newtcbtheorem{challange}{Challange}{challangestyle}{challange}

\title{LMTuner: An user-friendly and highly-integrable Training Framework \\for fine-tuning Large Language Models}
\author{
     Yixuan Weng $^1$, Zhiqi Wang$^{1,2}$, Huanxuan Liao$^{1,2}$, Shizhu He$^{1,2}$, Shengping Liu$^{3}$, Kang Liu$^{1,2}$, Jun Zhao$^{1,2}$ \\
  $^1$ The Laboratory of Cognition and Decision Intelligence for Complex Systems, \\Institute of Automation, Chinese Academy of Sciences \\
  $^2$School of Artificial Intelligence, University of Chinese Academy of Sciences\\
  $^3$Unisound, Beijing, China\\
  	{wengsyx@gmail.com}, {\{wangzhiqi2022, liaohuanxuan2023\}@ia.ac.cn}, \\
	liushengping@unisound.com, {\{shizhu.he, kliu, jzhao\}@nlpr.ia.ac.cn}
\\
}
\begin{document}
\maketitle
\begin{abstract}
With the burgeoning development in the realm of large language models (LLMs), the demand for efficient incremental training tailored to specific industries and domains continues to increase. Currently, the predominantly employed frameworks lack modular design, it often takes a lot of coding work to kickstart the training of LLM. To address this, we present "LMTuner", a highly usable, integrable, and scalable system for training LLMs expeditiously and with minimal user-input. LMTuner comprises three main modules - the Interaction, Training, and Inference Modules. We advocate that LMTuner's usability and integrality alleviate the complexities in training large language models. Remarkably, even a novice user could commence training large language models within five minutes. Furthermore, it integrates DeepSpeed frameworks and supports Efficient Fine-Tuning methodologies like Low Rank Adaptation (LoRA), Quantized LoRA (QLoRA), etc., enabling the training of language models scaling from 300M to a whopping 130B parameters using a single server. The LMTuner's homepage\footnote{\url{https://wengsyx.github.io/LMTuner/}}and screencast video \footnote{\url{https://youtu.be/nsXmWOmN3rE}}are now publicly available. 
\end{abstract}

\section{Introduction}

Large language models (LLMs) are demonstrating unprecedented performance in numerous natural language understanding and generation tasks \cite{wei2022chain,wei2022emergent,Weng2023LargeLM} thanks to their capacity to learn from extensive text data with generative manner~ \cite{brown2020language,scao2022bloom}. This has led to a rising number of researchers and engineers embarking on the training of their own language models for specific industries and domains \cite{alpaca,xu2023baize,cui2023chatlaw}. However, training LLMs imposes high demands on engineering skills \cite{zhang2022opt}, and the various techniques applicable to such training are mostly disparate \cite{dao2022flashattention,xi2023training,luo2023came}. This not only increases the complexity of related projects but also adds to the learning cost required for training language models.

\begin{figure}[t]
\begin{center}
	\includegraphics[width=0.47\textwidth]{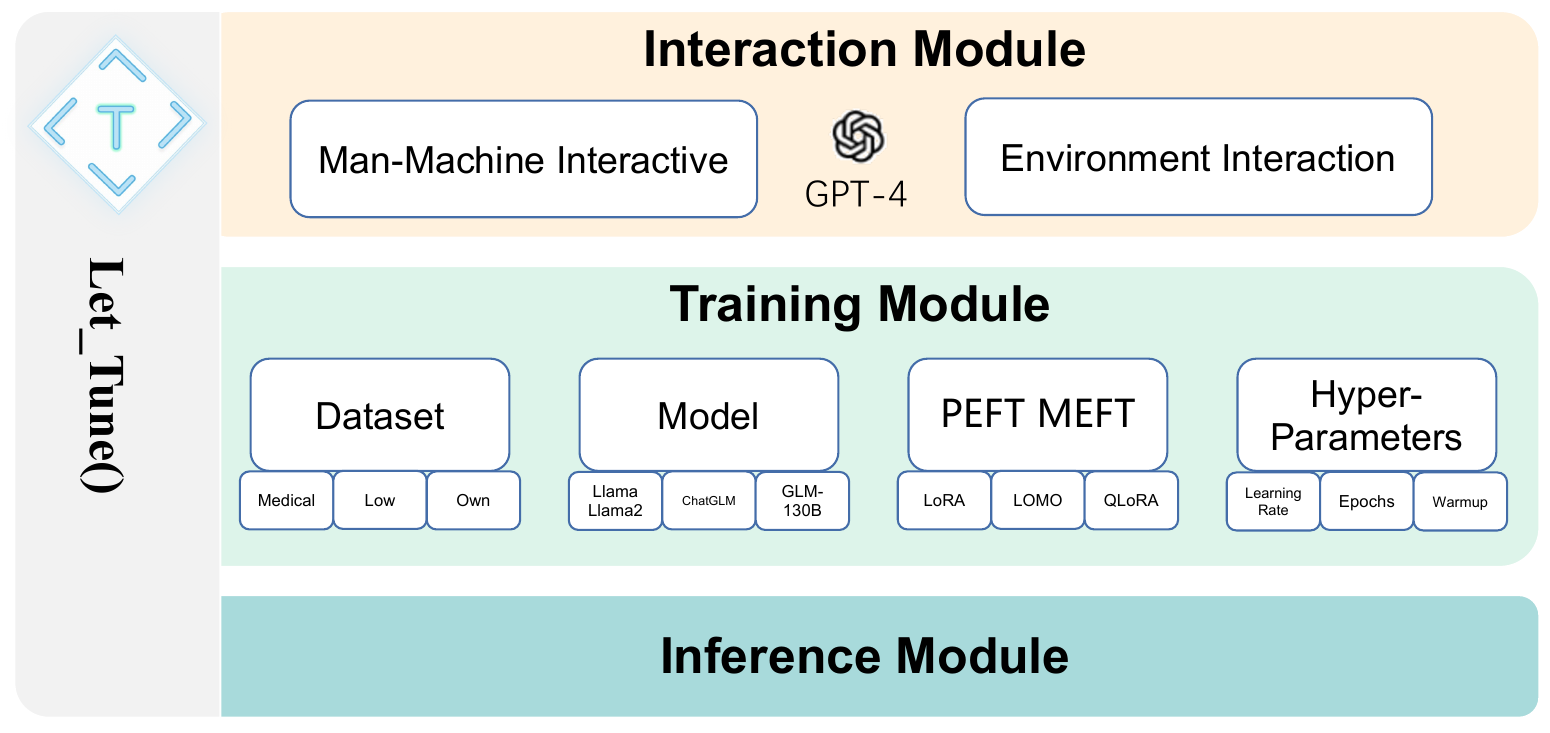}
\end{center}
\vspace{-0.3cm}
 \caption{The architecture of the LMTuner. LMTuner can be invoked through a single line of code \raisebox{-0.22\height}{\includegraphics[width=45pt]{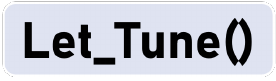}}. The overall process is top-down, sequentially going through the Interaction Module, Training Module, and Inference Module.}
\label{figure:1}
\vspace{-0.3cm}
\end{figure}

%%⬇️改动，替换下一段
% With the progressive development of generative large language model technology, various techniques have emerged, incorporated into different toolkits. Presently, when aiming for model parallelism, MegatronLM stands as a preferred choice, while bitsandbytes serves the purpose of model quantization, and opendelta facilitates the implementation of PEFT technology. Nonetheless, the flexibility offered by these individual tools comes at the cost of developers spending significant time coordinating the usage of diverse tool modules, rendering direct application challenging.

% To address this issue, several frameworks like h2oGPT and Lamini have attempted to integrate these functionalities. However, despite their existence, developers still need the extensive engineering and coding tasks that hinder full detachment from these complexities. The integration of multiple functionalities remains an arduous process, impeding the swift adoption and widespread dissemination of these cutting-edge techniques.

% To address these challenges, several frameworks, such as h2oGPT and lamini, have attempted to consolidate some of these functionalities. However, these frameworks typically integrate only a subset of the available techniques, often lacking comprehensive coverage of commonly used model technologies.

%%⬆️改动，替换下一段

With the progressive development of generative large language model technology, various techniques have emerged, incorporated into different toolkits \cite{zhao-etal-2023-tencentpretrain}. As shown in Table \ref{table:1}, when aiming for model parallelism, MegatronLM \cite{shoeybi2020megatronlm} stands as a preferred choice, while bitsandbytes \cite{dettmers2022llmint8} serves the purpose of model quantization, and Opendelta \cite{hu-etal-2023-opendelta} facilitates the implementation of Efficient Fine-Tuning technology. Nonetheless, the flexibility offered by these individual tools comes at the cost of developers spending significant time coordinating the usage of diverse tool modules, rendering direct application challenging. To address these challenges, several frameworks, such as h2oGPT \cite{candel2023h2ogpt} and Lamini \cite{Lamini2023}, have attempted to consolidate some of these functionalities. However, these frameworks typically integrate only a subset of the available techniques, often lacking comprehensive coverage of commonly used model technologies.

% As language models have recently become progressively homogeneous, the possibility of modular integration of training frameworks has emerged \cite{zhao-etal-2023-tencentpretrain}. Some related studies have attempted to provide users who are training models with modular functions \cite{ding-etal-2022-openprompt,hu-etal-2023-opendelta}, yet these efforts still fall short in term of user-friendliness and in adapting to training requirements for large language models. Sometimes, the time users spend learning frameworks and debugging code can even exceed the time required to train models. 

%%⬇️改动，替换下一段
% Therefore, in this paper, we present LMTuner, a novel framework that significantly minimizes these obstacles by offering an easy-to-use, scalable, and integrative modular system. We have integrated all the necessary tools for training a generative language model, while also providing developers with an interactive module. With this module, developers can effortlessly generate their desired models and automatically pass on the parameters to the next module for seamless training. This streamlines the entire process, allowing the system to be readily deployed and utilized upon completion of training.
%%⬆️改动，替换下一段

Therefore, in this paper, we present LMTuner, a novel framework that significantly minimizes these obstacles by offering an easy-to-use, scalable, and integrative modular system. As depicted in Figure \ref{figure:1}, LMTuner is comprised of three modules. 1) The Interaction Module enables user-friendly communication, automatically adjusting parameters based on user needs and context, ideal for non-technical users. 2) The Training Module autonomously processes training using these parameters, saving users the complexity of setup. 3) The Inference Module utilizes the trained models for various tasks upon training completion. With these modules, developers can effortlessly generate their desired models and automatically pass on the parameters to the next module for seamless training. This streamlines the entire process, allowing the system to be readily deployed and utilized upon completion of training.

In summary, our proposed LMTuner system offers greater usability and flexibility, allowing users to swiftly configure parameters for training language models according to specific needs and effectively initiate the training. Our contributions are as follows:

\begin{itemize}
\item We have proposed LMTuner, which is the highly usable and integrated training system for LLMs. It is free to use and license-friendly (Apache 2.0). And we open source code at \url{https://github.com/WENGSYX/LMTuner}.

\item LMTuner boasts high usability, only needing a single code (\raisebox{-0.25\height}{\includegraphics[width=50pt]{emnlp2023-latex/Let_Tune.pdf}}) to be launched. It facilitates quick start-up for large language model training by allowing users to interact using natural language with LMTuner.

\item We have incorporated a wide range of techniques suited for training large language models, including models, domain QA datasets, Efficient Fine-Tuning methods, and specific hyperparameters. This integration fosters the research and development of large language models.
\end{itemize}

\begin{table*}
\resizebox{\textwidth}{!}{%
\centering
\begin{tabular}{l|ccccccccc}

\toprule
& \multicolumn{7}{c}{\textbf{Highly-integrable}} & \multicolumn{2}{c}{\textbf{User-friendly}} \\
\cmidrule(lr){2-8} \cmidrule(l){9-10} 
 \space & Model Parallelism & Quantization & PEFT & MEFT & ZeRO & Load Dataset& Position Interpolation & AI Assisstent & Code Concise \\
\midrule
MegatronLM \cite{shoeybi2020megatronlm}& \checkmark & \space & \space & \space & \space & \space & \space & \space \\
Huggingface \cite{wolf-etal-2020-transformers}& \checkmark & \space & \checkmark & \space & \checkmark & \checkmark & \space & \space & \checkmark \\
bitsandbytes \cite{dettmers2022llmint8}& \space & \checkmark & \space & \space & \space & \space & \space & \space \\
OpenDelta \cite{hu-etal-2023-opendelta} & \space & \space & \checkmark & \space & \space & \space & \space & \space & \checkmark\\
Lamini \cite{Lamini2023} & \space & \space & \space& \space & \space & \checkmark & \space & \space & \checkmark\\
h2oGPT \cite{candel2023h2ogpt} & \space & \checkmark & \checkmark & \space & \space & \checkmark & \space& \space & \checkmark \\

\midrule
\textbf{LMTuner (Ours)} & \checkmark & \checkmark & \checkmark & \checkmark & \checkmark & \checkmark & \checkmark & \checkmark & \checkmark  \\

\bottomrule
\end{tabular}}
\vspace{-0.1cm}
\caption{Compared to other commonly used language model training systems, LMTuner system has highly integrable and user-friendly.}
\label{table:1}
\vspace{-0.15cm}
\end{table*}
\section{Related Work}

The development of pre-training language models has brought about numerous language model tools and a flourishing NLP community, among which are ``Transformers''  \cite{wolf-etal-2020-transformers}. It establishes a range of model classes and provides APIs for implementing easily extendable transformer models. Other tools closely associated with language model building include Fairseq \cite{ott-etal-2019-fairseq}, MegatronLM \cite{shoeybi2020megatronlm}, MedConQA \cite{xia-etal-2022-medconqa}, OpenDelta \cite{hu-etal-2023-opendelta} and h2oGPT \cite{candel2023h2ogpt}. Differing from these, LMTuner is specifically designed for training auto-regressive LLMs. With its modular design, it allows for a free combination of different pre-training models, datasets, model frameworks, automatic length extrapolation settings, and PEFT methods within one framework. With the assistance of dialogue-type LLMs such as GPT-4 \cite{openai2023gpt4}, users can accomplish the entire training process with just a single line of code. 

Recent researches have proposed many directions for training LLMs \cite{ignat2023phd}, including high-quality data \cite{zhou2023lima,gunasekar2023textbooks}, efficient fine-tuning methods \cite{peft}, model structures \cite{shazeer2019fast,dao2022flashattention}, and length extrapolation \cite{chi2023dissecting,tworkowski2023focused,su2022roformer,chen2023extending}. The LMTuner system integrates the most advanced techniques in these fields, allowing users to make choices according to their needs. LMTuner integrates these techniques into separate modules, facilitating user selection and usage \cite{wang2023zero,dao2022flashattention,dettmers2023qlora}.

\begin{figure*}[t]
\begin{center}
	\includegraphics[width=\textwidth]{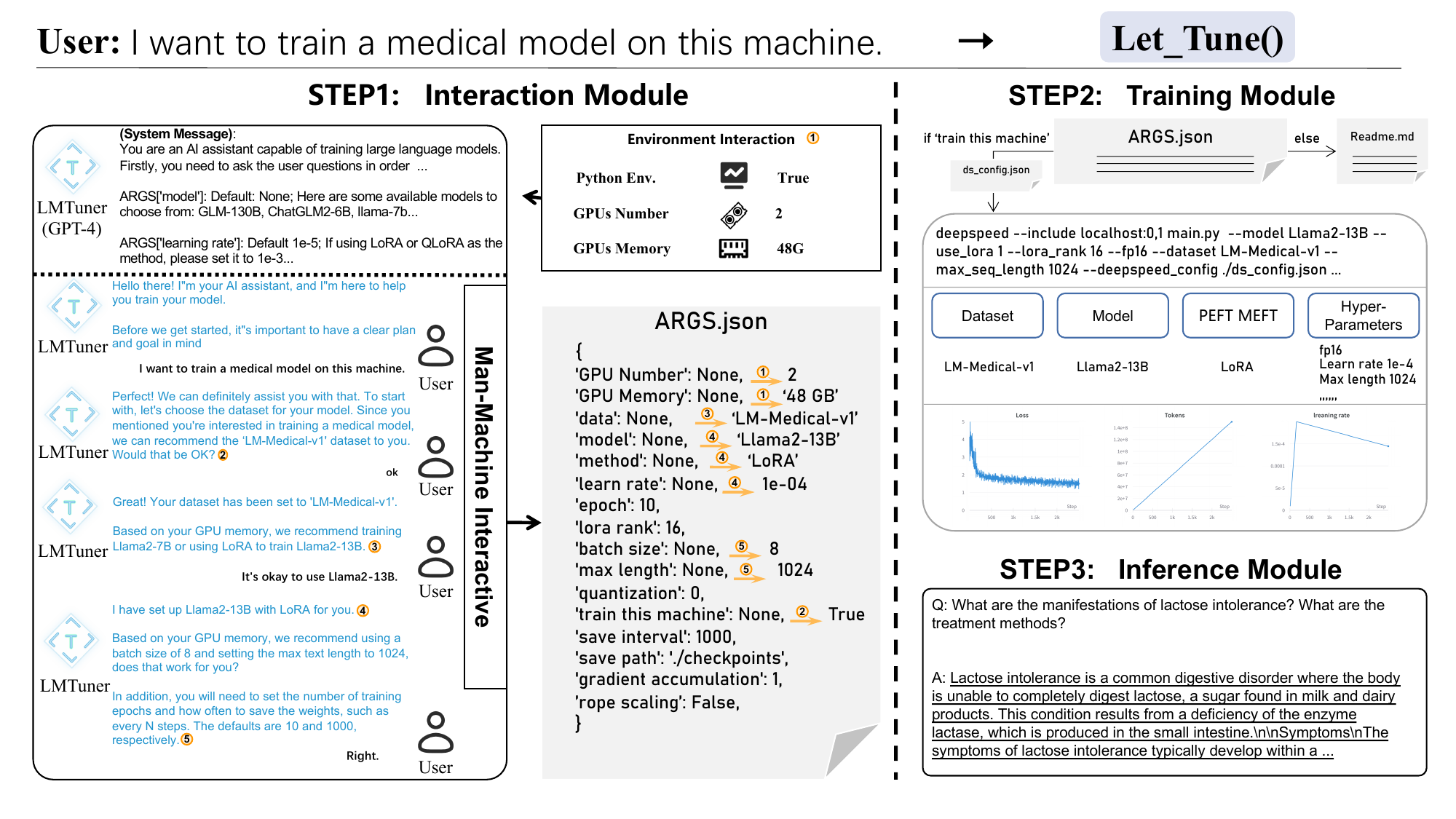}
\end{center}
\vspace{-0.15cm}
 \caption{Overview of the LMTuner system. When a user makes a request by \raisebox{-0.23\height}{\includegraphics[width=45pt]{emnlp2023-latex/Let_Tune.pdf}}, the LMTuner system automatically detects the GPUs available in the environment and engages in detailed discussions with the user to determine the configuration. The configuration is then automatically sent to the Training Module to initiate training. During the training process, wandb is enabled to save training logs. The final trained model is sent to the Inference Module for deployment.}
\label{figure:2}
\vspace{-0.1cm}
\end{figure*}

LLMs undergoing instruction pre-training can align with human instructions \cite{ouyang2022training,chung2022scaling} and have been discovered to possess many capabilities not present in smaller language models \cite{zhao2023survey}, such as tool creation \cite{cai2023large}, environment exploration \cite{wang2023voyager}, self-verification \cite{weng2022large}, and complex reasoning \cite{weng2023neural,zhu2023towards}. Leveraging these capabilities of LLMs, LMTuner's Interaction Module can help users analyze their needs and recommend necessary settings through a nature-language-based interaction method. This practice of using LLMs as interactive agents \cite{wang2023interactive} have also been applied in interactive decision making \cite{yao2023react} and functioning as research assistants \cite{ren2023crest}.

\section{LMTuner}

LMTuner is an open-source system that offers a command-line interface (CLI) for training LLMs:

\begin{figure}[!thp]
\centering
\begin{minipage}{0.96\linewidth}
\begin{lstlisting}[language=Python]
#Official Launch for LMTuner.
from LMTuner import Let_Tune
Let_Tune()

>>> [AI] Greetings! I am your AI assistant. Present assist training your model. Necessary possess clear plan, goal first.
>>> [ANSWER] :

\end{lstlisting}
\end{minipage}
\label{fig:code}
\vspace{-0.1cm}
\end{figure}

As shown in Figure \ref{figure:2}, LMTuner allows for the development of LLMs training through simple conversations. This can improve engineering efficiency when training LLMs and reduce code burden.

\subsection{Interaction Module}

The Interaction Module of LMTuner, utilizing GPT-4's (or ChatGPT) System Message and Function features, serves as an LLM training assistant. This module, during initialization, incorporates common training issues, parameter configurations, and selectable methods into System Messages, thereby streamlining the training process. The essential function of GPT-4 within this context is to ascertain the requisite training configurations. A function, Set\_ARGS, is available to amend the parameters in the training configuration. To enable GPT-4 to evaluate various parameter settings, we utilize pynvml to monitor server GPUs, integrating this information within the System Message content.

The Interaction Module helps prevent user configuration mistakes through its user-friendly interface and high adaptability. As shown in Figure 3, users can state their needs conversationally. LMTuner then analyzes their words and suggests appropriate training settings. This works because the System Message contains knowledge about LLM training, compensating for the common lack of such expertise among researchers and engineers without backgrounds in large language model training.
\begin{figure*}[t]
\begin{center}
	\includegraphics[width=\textwidth]{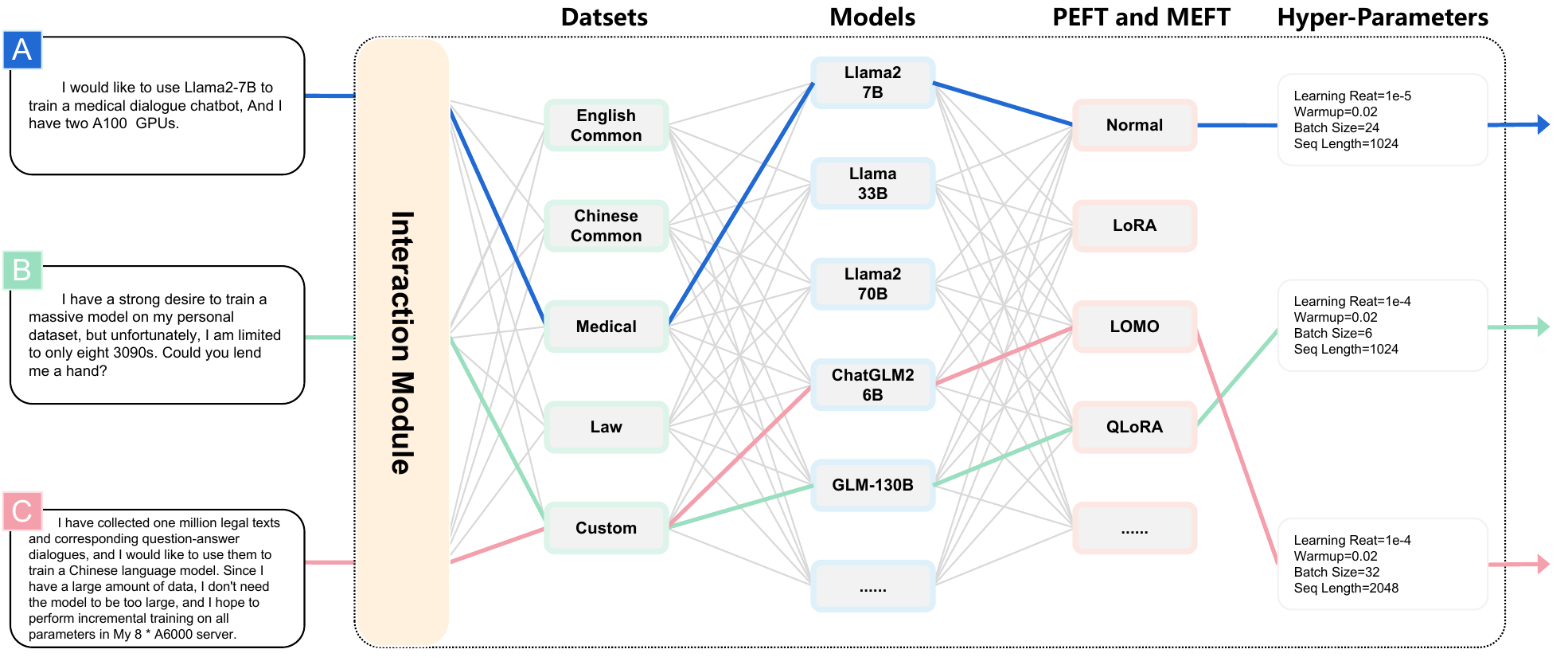}
\end{center}
\vspace{-0.1cm}
 \caption{Depending on the requirements specified, LMTuner can automatically suggest different training plans.}
\label{figure:3}
\vspace{-0.1cm}
\end{figure*}

The Interaction Module, while offering immense flexibility and ease of use, prevents potential issues such as user configuration errors. As shown in Figure \ref{figure:3}, users with different requirements only need to express their needs in natural language. The LMTuner system is capable of analyzing and recommending appropriate training parameters. If the user is not inclined to train the model on the current device, LMTuner will automatically generate a \texttt{Readme.md} file that includes environment configuration, model processing, and training code instructions like Figure \ref{figure:4}, facilitating a swift setup for the user on a new device.

Once all training parameters are finalized, LMTuner saves a copy of these in an \texttt{ARGS.json} file. If one wishes to initiate training quickly with the same parameters, they only need to pass the path name of \texttt{ARGS.json} to the \raisebox{-0.25\height}{\includegraphics[width=50pt]{emnlp2023-latex/Let_Tune.pdf}} function, thereby avoiding redundant repeated dialogues.

\begin{figure}[!thp]
\centering
\begin{minipage}{0.93\linewidth}
\begin{lstlisting}[language=Python]
#Quickly Launch for LMTuner.
from LMTuner import Let_Tune
Let_Tune(ARGS='./ARGS.config')

>>> [LMTuner] We will train the model~ Go!
>>> [2023-07-19 05:18:34,778] [INFO] [runner.py:555:main] cmd = python -u -m deepspeed.launcher.launch --world_info=xxxxx main.py --seed 1234 ......

\end{lstlisting}
\end{minipage}
\vspace{-0.2cm}
\label{fig:code}
\end{figure}

\begin{figure}[t]
\begin{center}
	\includegraphics[width=0.465\textwidth]{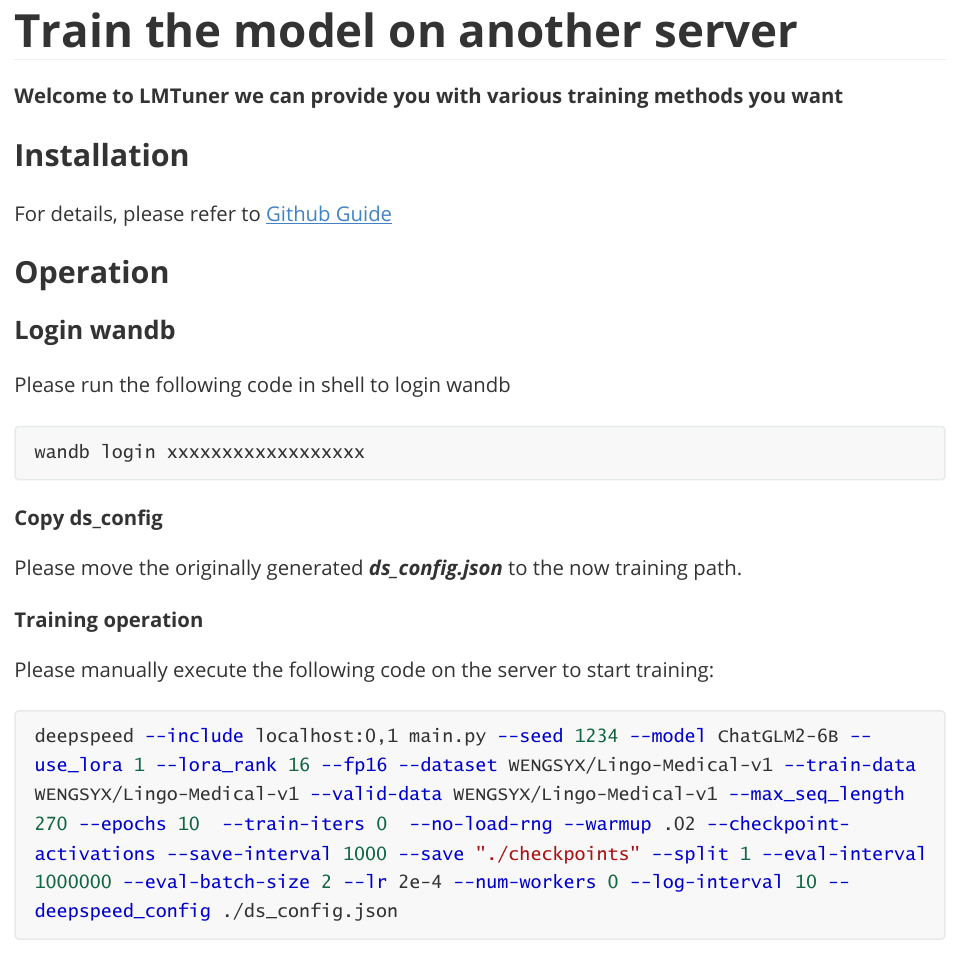}
\end{center}
\vspace{-0.2cm}
 \caption{An example of the \texttt{Readme.md} automatically generated by LMTuner.}
 \vspace{-0.1cm}
\label{figure:4}
\end{figure}

\subsection{Training Module}

The training module in LMTuner has highly integrated, easy to invoke, and extensible features. Currently, the mainstream LLMs are mostly similar in architecture, and the training process and loss calculation are generally consistent. Therefore, we construct the required techniques in a modularized manner at the code level according to different requirements. Meanwhile, such design also facilitates engineers with coding skills to directly replace the corresponding modules through hooks. In the rest of this section, we will introduce each technical module separately.

\noindent\textbf{Datasets.}
The availability of high-quality training data is crucial for developing capable LLMs. To facilitate affordable access to suitable datasets for question answering, we have curated and prepared a collection of QA datasets covering diverse domains, including English, Chinese, medical, and legal fields. Furthermore, to enable the development of models with customizable names and personas, the datasets have been augmented with synthetic question-answer pairs inquiring about the model's identity (e.g. "Hi~ I'm [MODEL NAME]"). During training, the [MODEL NAME] tokens can be dynamically substituted with the preferred name for each model instance.

\begin{figure}[!thp]
\centering
\begin{minipage}{0.97\linewidth}
\begin{lstlisting}[language=Python]
from LMTuner.dataset import LMTunerDataset

dataset = LMTunerDataset()

# Give your model a name
dataset.set_model_name('LMTuner')

# Add QA dataset samples
dataset.add_sample(['Who are you?',
                    "I'm LMTuner,  your personal sidekick!"])
\end{lstlisting}
\end{minipage}
\vspace{-0.1cm}
\label{fig:code}
\end{figure}

However, despite the richness of our curated datasets, they inherently possess certain limitations in coverage and diversity. To augment the built-in datasets and account for user-specific needs, our system also provides seamless support for customized training data. Users can simply provide the local path to their own JSONL-formatted question answering data files. This design choice provides greater flexibility to users, empowering them to tailor the training distribution to their unique application requirements. For instance, users can provide proprietary datasets containing sensitive or confidential information not suitable for public release. The ability to directly use local JSONL files avoids the need to port datasets to external platforms. 
\lstset{escapeinside={<@}{@>}}
\definecolor{ao}{rgb}{0.0, 0.5, 0.0}
\definecolor{byzantine}{rgb}{0.5, 0.2, 0.64}

\begin{lstlisting}[basicstyle=\small]
<@\textcolor{ao}{\# Pretrained Custom-Dataset Format}@>
{
  <@\textcolor{byzantine}{"input"}@>: "",
  <@\textcolor{byzantine}{"output"}@>: "With the burgeoning development in the realm of ...",
}

<@\textcolor{ao}{\# Instruct Custom-Dataset Format}@>
{
  <@\textcolor{byzantine}{"input"}@>: "Human: Who are you?",
  <@\textcolor{byzantine}{"output"}@>: " Assistant: I'm LMTuner,  your personal sidekick!",
}

\end{lstlisting}

\noindent\textbf{Models.}
Recent advancements in natural language processing have been enabled by the transformer architecture \cite{vaswani2017attention}. The SwissArmyTransformer \footnote{\url{https://github.com/THUDM/SwissArmyTransformer}} framework facilitates efficient development of diverse transformer models by decoupling reusable core components from interchangeable model-specific modules. These lightweight modules attach to the shared backbone via hooks, enabling rapid iteration and customization. In contrast, Transformers \cite{wolf-etal-2020-transformers} provides optimized implementations of canonical architectures and pretrained models for production use.

The LMTuner system combines these complementary strengths for flexible model development and deployment. It utilizes SwissArmyTransformer to construct tailored architectures and seamlessly integrates Transformers' pretrained models. This synthesis of code modularization and extensive pretrained models promises to enhance productivity, accelerate innovation, and improve real-world language understanding. LMTuner promotes exploratory modeling by innovating new designs built on transformer infrastructure, while benefiting from cutting-edge advancements in language model pretraining. Selectively utilizing both libraries stands to meaningfully advance natural language processing systems through rapid prototyping of specialized models and accessible deployment of state-of-the-art capabilities.

\noindent\textbf{Efficient Fine-Tuning}. LMTuner uses ZeRO technology of Deepspeed by default to unload parameters and improve training throughput. In addition, LMTuner provides parameters-efficient fine-tuning (PEFT) methods including LoRA \cite{hu2022lora} and QLoRA \cite{dettmers2023qlora}, and memory-efficient fine-tuning (MEFT) methods including LOMO \cite{Lv2023FullPF} and Quantization\cite{gholami2021survey}. They support training LLMs with low memory usage. These methods are implemented in a modularized manner at the code level of LMTuner, so they can be easily combined and used freely. 

\noindent\textbf{Position Interpolation.} To better support long-context modeling, LMTuner has integrated some scaling of RoPE. We implemented Xpos \cite{sun2022lengthextrapolatable} and some recent position interpolation methods like linear interpolation \cite{chen2023extending}, dynamic interpolation, NTK-Aware Scaled RoPE (NTKv1) and NTK-By-Parts (NTKv2). The dynamic methods choose the correct \textit{scale} parameter based on sequence length, rather than having to settle for a fixed trade-off between maximum sequence length and performance on shorter sequences, i.e., use the exact position values for the first 2048 contexts and then recalculate the position vectors for each new sequence length as the model generates the markers one by one. A certain degree of long-context modeling can be achieved by choosing different scaling methods.

\noindent\textbf{Other Details.} We use the probability distribution over sequences of tokens as the optimization objective with cross-entropy loss \cite{Radford2019LanguageMA} and use Lion optimizer \cite{https://doi.org/10.48550/arxiv.2302.06675} by default to optimize LLMs, because it has been proven to be more memory-efficient than Adam \cite{kingma2017adam}. During training, we record the loss, learning rate, and number of tokens respectively at each step using wandb\footnote{\url{https://wandb.ai/}}, and display them in the browser through line charts, which helps users observe the training status during training.

\subsection{Inference Module}

LMTuner loads the final model weights after training and generates continuations conditioned on given contexts until reaching the maximum length. To speed up inference, LMTuner provides model quantization methods including INT8 and INT4 quantization \cite{zeng2022glm}. By quantizing 16-bit floating-point weights into lower bitwidth integers such as 8-bit or 4-bit, the computation time and memory usage during inference can be reduced. LMTuner quantizes weights of selected layers in a trained model, while keeping activations in 16-bit floating-point format. After quantization, the inference latency on CPU and throughput on GPU can be improved significantly with little degradation in model quality \cite{liu2023emergent}.

\section{A Running Case}
The target audience of LMTuner is machine learning engineers and researchers across academia and industry. Novice users can leverage LMTuner's guided interaction while experts retain full control over implementation details. It compares favorably to current systems by combining user-friendliness, scalability, and integrability within a unified interface. LMTuner is open-source under the Apache 2.0 license, allowing free use in commercial products. By open-sourcing LMTuner, we hope to catalyze progress in large language model training and lower the barriers to leveraging these transformative technologies. The availability of an easy-to-use, highly customizable system should benefit the broader community.

Assuming we need to train a medical LLM that can assist in patient diagnosis, and we have two A6000 GPUs with 48GB VRAM each, as well as a medical QA dataset (MedDialog \cite{zeng2020meddialog}). Using LMTuner, we can automatically determine the training process, including the selection of the LLama-7B model and a set of corresponding hyperparameters\footnote{We released the code at \url{https://github.com/WENGSYX/LMTuner/tree/main/Example/English_Medical}}.

%Figure 2 displays the logged information during the training process,
while Table \ref{table:english} showcases the performance of the MedDialog test set (without manual model selection). Instead, we chose the final model obtained after LMTuner training, which completed 10 epochs. We observed that the LMTuner-trained model surpasses existing models in metrics such as Bleu, Meteor, and NIST. This indicates that the trained model is ready for direct utilization.

\begin{table}[h]
\renewcommand\arraystretch{1.3}
\resizebox{\linewidth}{!}{%
	\centering \small
   \begin{tabular}{c|rrrrr}

    \bottomrule \bottomrule
    
     \multicolumn{1}{c|}{\multirow{2}{*}{\textbf{Method}}}                                                                                  & \multicolumn{5}{c}{\textbf{MedDialog}}           \\                                             \multicolumn{1}{c|}{}&BLEU-2& BLEU-4&Meteor&Nist-2&Nist-4      \\ \hline

       GPT-3 (175B) + Direct&       14.93&7.52&4.55&0.814&0.852       \\
     GPT-3 (175B) + CoT & 9.18&4.61&3.54&0.546&0.569             \\ 

       Instruct-GPT (175B) + Direct&15.93&8.21&5.74&0.874&0.913              \\
     Instruct-GPT (175B) + CoT &16.49&7.80&4.94&1.020&1.059                \\ 

            GLM (130B) + Direct&12.16&6.02&5.31&0.577&0.601              \\
     GLM (130B) + CoT &30.02&15.61&8.73&1.909&2.017            \\ \hline \multicolumn{1}{c|}{\citet{zeng2020meddialog}} &31.67&16.88&9.57&1.981&2.076                                                                        \\ 
     \textbf{LMTuner (Ours)}&\textbf{35.62}&\textbf{19.02}&\textbf{9.47}& \textbf{2.377}&\textbf{2.517}\\

 \bottomrule  \bottomrule
        
    \end{tabular}}
    
    \caption{We present the results of several language models on the MedDialog task, including the zero-shot (Direct) and CoT \cite{wei2022chain,weifinetuned} performance of some LLMs (GPT-3:\texttt{code-davinci-001} \cite{DBLP:journals/corr/abs-2107-03374}; Instruct-GPT: \texttt{code-davinci-002} \cite{LongOuyang2022TrainingLM}), as well as the performance of previous Finetune state-of-the-art models.}
	\label{table:english}
\vspace{-0.1cm}
\end{table}

\section{Conclusion}
LMTuner represents a pioneering effort to facilitate large language model training through enhanced usability and modularity. We believe LMTuner represents a significant step towards realizing the full potential of large language models. By continuing to integrate emerging techniques and community feedback, its capabilities will only grow over time. The availability of LMTuner as an open source project presents exciting opportunities for LLMs enhancement. We hope its emphasis on usability and extensibility will meaningfully accelerate future work in this paradigm-defining domain. 

\section*{Limitations}
While LMTuner is designed to be user-friendly and intuitive, it may not capture all specific user requirements in its current version. The training process might need to be adjusted or additional techniques might need to be integrated to achieve optimal performance for certain specialized tasks. Some complex requirements could necessitate manual code modification, which might increase the learning curve for non-expert users. However, this limitation is also being actively addressed through continuous development and updates to the system to enhance its comprehension of user requirements.

\section*{Acknowledgments}
This work was supported by the Strategic Priority Research Program of Chinese Academy of Sciences (No. XDA27020100) and the National Natural Science Foundation of China (No.U1936207,  No.61976211). This work was supported by the Youth Innovation Promotion Association CAS and Yunnan Provincial Major Science and Technology Special Plan Projects (No.202202AD080004).
\bibliography{anthology,custom}

\begin{thebibliography}{63}
\expandafter\ifx\csname natexlab\endcsname\relax\def\natexlab#1{#1}\fi

\bibitem[{Black et~al.(2021)Black, Leo, Wang, Leahy, and Biderman}]{gpt-neo}
Sid Black, Gao Leo, Phil Wang, Connor Leahy, and Stella Biderman. 2021.
\newblock \href {https://doi.org/10.5281/zenodo.5297715} {{GPT-Neo: Large Scale
  Autoregressive Language Modeling with Mesh-Tensorflow}}.
\newblock {If you use this software, please cite it using these metadata.}

\bibitem[{Brown et~al.(2020)Brown, Mann, Ryder, Subbiah, Kaplan, Dhariwal,
  Neelakantan, Shyam, Sastry, Askell et~al.}]{brown2020language}
Tom Brown, Benjamin Mann, Nick Ryder, Melanie Subbiah, Jared~D Kaplan, Prafulla
  Dhariwal, Arvind Neelakantan, Pranav Shyam, Girish Sastry, Amanda Askell,
  et~al. 2020.
\newblock Language models are few-shot learners.
\newblock \emph{Advances in neural information processing systems},
  33:1877--1901.

\bibitem[{Cai et~al.(2023)Cai, Wang, Ma, Chen, and Zhou}]{cai2023large}
Tianle Cai, Xuezhi Wang, Tengyu Ma, Xinyun Chen, and Denny Zhou. 2023.
\newblock \href {http://arxiv.org/abs/2305.17126} {Large language models as
  tool makers}.

\bibitem[{Candel et~al.(2023)Candel, McKinney, Singer, Pfeiffer, Jeblick,
  Prabhu, Gambera, Landry, Bansal, Chesler, Lee, Conde, Stetsenko, Grellier,
  and Ambati}]{candel2023h2ogpt}
Arno Candel, Jon McKinney, Philipp Singer, Pascal Pfeiffer, Maximilian Jeblick,
  Prithvi Prabhu, Jeff Gambera, Mark Landry, Shivam Bansal, Ryan Chesler,
  Chun~Ming Lee, Marcos~V. Conde, Pasha Stetsenko, Olivier Grellier, and
  SriSatish Ambati. 2023.
\newblock \href {http://arxiv.org/abs/2306.08161} {h2ogpt: Democratizing large
  language models}.

\bibitem[{Chen et~al.(2021)Chen, Tworek, Jun, Yuan, de~Oliveira~Pinto, Kaplan,
  Edwards, Burda, Joseph, Brockman, Ray, Puri, Krueger, Petrov, Khlaaf, Sastry,
  Mishkin, Chan, Gray, Ryder, Pavlov, Power, Kaiser, Bavarian, Winter, Tillet,
  Such, Cummings, Plappert, Chantzis, Barnes, Herbert{-}Voss, Guss, Nichol,
  Paino, Tezak, Tang, Babuschkin, Balaji, Jain, Saunders, Hesse, Carr, Leike,
  Achiam, Misra, Morikawa, Radford, Knight, Brundage, Murati, Mayer, Welinder,
  McGrew, Amodei, McCandlish, Sutskever, and
  Zaremba}]{DBLP:journals/corr/abs-2107-03374}
Mark Chen, Jerry Tworek, Heewoo Jun, Qiming Yuan, Henrique~Ponde
  de~Oliveira~Pinto, Jared Kaplan, Harrison Edwards, Yuri Burda, Nicholas
  Joseph, Greg Brockman, Alex Ray, Raul Puri, Gretchen Krueger, Michael Petrov,
  Heidy Khlaaf, Girish Sastry, Pamela Mishkin, Brooke Chan, Scott Gray, Nick
  Ryder, Mikhail Pavlov, Alethea Power, Lukasz Kaiser, Mohammad Bavarian,
  Clemens Winter, Philippe Tillet, Felipe~Petroski Such, Dave Cummings,
  Matthias Plappert, Fotios Chantzis, Elizabeth Barnes, Ariel Herbert{-}Voss,
  William~Hebgen Guss, Alex Nichol, Alex Paino, Nikolas Tezak, Jie Tang, Igor
  Babuschkin, Suchir Balaji, Shantanu Jain, William Saunders, Christopher
  Hesse, Andrew~N. Carr, Jan Leike, Joshua Achiam, Vedant Misra, Evan Morikawa,
  Alec Radford, Matthew Knight, Miles Brundage, Mira Murati, Katie Mayer, Peter
  Welinder, Bob McGrew, Dario Amodei, Sam McCandlish, Ilya Sutskever, and
  Wojciech Zaremba. 2021.
\newblock \href {http://arxiv.org/abs/2107.03374} {Evaluating large language
  models trained on code}.
\newblock \emph{CoRR}, abs/2107.03374.

\bibitem[{Chen et~al.(2023{\natexlab{a}})Chen, Wong, Chen, and
  Tian}]{chen2023extending}
Shouyuan Chen, Sherman Wong, Liangjian Chen, and Yuandong Tian.
  2023{\natexlab{a}}.
\newblock \href {http://arxiv.org/abs/2306.15595} {Extending context window of
  large language models via positional interpolation}.

\bibitem[{Chen et~al.(2023{\natexlab{b}})Chen, Liang, Huang, Real, Wang, Liu,
  Pham, Dong, Luong, Hsieh, Lu, and
  Le}]{https://doi.org/10.48550/arxiv.2302.06675}
Xiangning Chen, Chen Liang, Da~Huang, Esteban Real, Kaiyuan Wang, Yao Liu, Hieu
  Pham, Xuanyi Dong, Thang Luong, Cho-Jui Hsieh, Yifeng Lu, and Quoc~V. Le.
  2023{\natexlab{b}}.
\newblock \href {https://arxiv.org/abs/2302.06675} {Symbolic discovery of
  optimization algorithms}.

\bibitem[{Chi et~al.(2023)Chi, Fan, Rudnicky, and Ramadge}]{chi2023dissecting}
Ta-Chung Chi, Ting-Han Fan, Alexander~I. Rudnicky, and Peter~J. Ramadge. 2023.
\newblock \href {http://arxiv.org/abs/2212.10356} {Dissecting transformer
  length extrapolation via the lens of receptive field analysis}.

\bibitem[{Chung et~al.(2022)Chung, Hou, Longpre, Zoph, Tay, Fedus, Li, Wang,
  Dehghani, Brahma, Webson, Gu, Dai, Suzgun, Chen, Chowdhery, Castro-Ros,
  Pellat, Robinson, Valter, Narang, Mishra, Yu, Zhao, Huang, Dai, Yu, Petrov,
  Chi, Dean, Devlin, Roberts, Zhou, Le, and Wei}]{chung2022scaling}
Hyung~Won Chung, Le~Hou, Shayne Longpre, Barret Zoph, Yi~Tay, William Fedus,
  Yunxuan Li, Xuezhi Wang, Mostafa Dehghani, Siddhartha Brahma, Albert Webson,
  Shixiang~Shane Gu, Zhuyun Dai, Mirac Suzgun, Xinyun Chen, Aakanksha
  Chowdhery, Alex Castro-Ros, Marie Pellat, Kevin Robinson, Dasha Valter,
  Sharan Narang, Gaurav Mishra, Adams Yu, Vincent Zhao, Yanping Huang, Andrew
  Dai, Hongkun Yu, Slav Petrov, Ed~H. Chi, Jeff Dean, Jacob Devlin, Adam
  Roberts, Denny Zhou, Quoc~V. Le, and Jason Wei. 2022.
\newblock \href {http://arxiv.org/abs/2210.11416} {Scaling
  instruction-finetuned language models}.

\bibitem[{Cui et~al.(2023)Cui, Li, Yan, Chen, and Yuan}]{cui2023chatlaw}
Jiaxi Cui, Zongjian Li, Yang Yan, Bohua Chen, and Li~Yuan. 2023.
\newblock \href {http://arxiv.org/abs/2306.16092} {Chatlaw: Open-source legal
  large language model with integrated external knowledge bases}.

\bibitem[{Dao et~al.(2022)Dao, Fu, Ermon, Rudra, and
  Ré}]{dao2022flashattention}
Tri Dao, Daniel~Y. Fu, Stefano Ermon, Atri Rudra, and Christopher Ré. 2022.
\newblock \href {http://arxiv.org/abs/2205.14135} {Flashattention: Fast and
  memory-efficient exact attention with io-awareness}.

\bibitem[{Dettmers et~al.(2022)Dettmers, Lewis, Belkada, and
  Zettlemoyer}]{dettmers2022llmint8}
Tim Dettmers, Mike Lewis, Younes Belkada, and Luke Zettlemoyer. 2022.
\newblock \href {http://arxiv.org/abs/2208.07339} {Llm.int8(): 8-bit matrix
  multiplication for transformers at scale}.

\bibitem[{Dettmers et~al.(2023)Dettmers, Pagnoni, Holtzman, and
  Zettlemoyer}]{dettmers2023qlora}
Tim Dettmers, Artidoro Pagnoni, Ari Holtzman, and Luke Zettlemoyer. 2023.
\newblock \href {http://arxiv.org/abs/2305.14314} {Qlora: Efficient finetuning
  of quantized llms}.

\bibitem[{Diamos et~al.(2023)Diamos, Zhou, Ibraheem, Daniel, and
  Arman}]{Lamini2023}
Greg Diamos, Sharon Zhou, Samee Ibraheem, Daniel, and Arman. 2023.
\newblock Lamini: The llm engine for rapidly customizing models.
\newblock \url{https://github.com/lamini-ai/lamini}.

\bibitem[{Du et~al.(2022)Du, Qian, Liu, Ding, Qiu, Yang, and Tang}]{du2022glm}
Zhengxiao Du, Yujie Qian, Xiao Liu, Ming Ding, Jiezhong Qiu, Zhilin Yang, and
  Jie Tang. 2022.
\newblock Glm: General language model pretraining with autoregressive blank
  infilling.
\newblock In \emph{Proceedings of the 60th Annual Meeting of the Association
  for Computational Linguistics (Volume 1: Long Papers)}, pages 320--335.

\bibitem[{Gholami et~al.(2021)Gholami, Kim, Dong, Yao, Mahoney, and
  Keutzer}]{gholami2021survey}
Amir Gholami, Sehoon Kim, Zhen Dong, Zhewei Yao, Michael~W. Mahoney, and Kurt
  Keutzer. 2021.
\newblock \href {http://arxiv.org/abs/2103.13630} {A survey of quantization
  methods for efficient neural network inference}.

\bibitem[{Gunasekar et~al.(2023)Gunasekar, Zhang, Aneja, Mendes, Giorno, Gopi,
  Javaheripi, Kauffmann, de~Rosa, Saarikivi, Salim, Shah, Behl, Wang, Bubeck,
  Eldan, Kalai, Lee, and Li}]{gunasekar2023textbooks}
Suriya Gunasekar, Yi~Zhang, Jyoti Aneja, Caio César~Teodoro Mendes, Allie~Del
  Giorno, Sivakanth Gopi, Mojan Javaheripi, Piero Kauffmann, Gustavo de~Rosa,
  Olli Saarikivi, Adil Salim, Shital Shah, Harkirat~Singh Behl, Xin Wang,
  Sébastien Bubeck, Ronen Eldan, Adam~Tauman Kalai, Yin~Tat Lee, and Yuanzhi
  Li. 2023.
\newblock \href {http://arxiv.org/abs/2306.11644} {Textbooks are all you need}.

\bibitem[{Hu et~al.(2022)Hu, Shen, Wallis, Allen-Zhu, Li, Wang, Wang, and
  Chen}]{hu2022lora}
Edward~J Hu, Yelong Shen, Phillip Wallis, Zeyuan Allen-Zhu, Yuanzhi Li, Shean
  Wang, Lu~Wang, and Weizhu Chen. 2022.
\newblock \href {https://openreview.net/forum?id=nZeVKeeFYf9} {Lo{RA}: Low-rank
  adaptation of large language models}.
\newblock In \emph{International Conference on Learning Representations}.

\bibitem[{Hu et~al.(2023)Hu, Ding, Zhao, Lv, Zhang, Liu, and
  Sun}]{hu-etal-2023-opendelta}
Shengding Hu, Ning Ding, Weilin Zhao, Xingtai Lv, Zhen Zhang, Zhiyuan Liu, and
  Maosong Sun. 2023.
\newblock \href {https://aclanthology.org/2023.acl-demo.26} {{O}pen{D}elta: A
  plug-and-play library for parameter-efficient adaptation of pre-trained
  models}.
\newblock In \emph{Proceedings of the 61st Annual Meeting of the Association
  for Computational Linguistics (Volume 3: System Demonstrations)}, pages
  274--281, Toronto, Canada. Association for Computational Linguistics.

\bibitem[{Ignat et~al.(2023)Ignat, Jin, Abzaliev, Biester, Castro, Deng, Gao,
  Gunal, He, Kazemi, Khalifa, Koh, Lee, Liu, Min, Mori, Nwatu, Perez-Rosas,
  Shen, Wang, Wu, and Mihalcea}]{ignat2023phd}
Oana Ignat, Zhijing Jin, Artem Abzaliev, Laura Biester, Santiago Castro, Naihao
  Deng, Xinyi Gao, Aylin Gunal, Jacky He, Ashkan Kazemi, Muhammad Khalifa,
  Namho Koh, Andrew Lee, Siyang Liu, Do~June Min, Shinka Mori, Joan Nwatu,
  Veronica Perez-Rosas, Siqi Shen, Zekun Wang, Winston Wu, and Rada Mihalcea.
  2023.
\newblock \href {http://arxiv.org/abs/2305.12544} {A phd student's perspective
  on research in nlp in the era of very large language models}.

\bibitem[{Kingma and Ba(2017)}]{kingma2017adam}
Diederik~P. Kingma and Jimmy Ba. 2017.
\newblock \href {http://arxiv.org/abs/1412.6980} {Adam: A method for stochastic
  optimization}.

\bibitem[{Liu et~al.(2023)Liu, Liu, Gao, Gao, Zhao, Li, Ding, and
  Wen}]{liu2023emergent}
Peiyu Liu, Zikang Liu, Ze-Feng Gao, Dawei Gao, Wayne~Xin Zhao, Yaliang Li,
  Bolin Ding, and Ji-Rong Wen. 2023.
\newblock \href {http://arxiv.org/abs/2307.08072} {Do emergent abilities exist
  in quantized large language models: An empirical study}.

\bibitem[{Luo et~al.(2023)Luo, Ren, Zheng, Jiang, Jiang, and You}]{luo2023came}
Yang Luo, Xiaozhe Ren, Zangwei Zheng, Zhuo Jiang, Xin Jiang, and Yang You.
  2023.
\newblock \href {http://arxiv.org/abs/2307.02047} {Came: Confidence-guided
  adaptive memory efficient optimization}.

\bibitem[{Lv et~al.(2023)Lv, Yang, Liu, jie Gao, Guo, and Qiu}]{Lv2023FullPF}
Kai Lv, Yuqing Yang, Tengxiao Liu, Qi~jie Gao, Qipeng Guo, and Xipeng Qiu.
  2023.
\newblock Full parameter fine-tuning for large language models with limited
  resources.

\bibitem[{Mangrulkar et~al.(2022)Mangrulkar, Gugger, Debut, Belkada, and
  Paul}]{peft}
Sourab Mangrulkar, Sylvain Gugger, Lysandre Debut, Younes Belkada, and Sayak
  Paul. 2022.
\newblock Peft: State-of-the-art parameter-efficient fine-tuning methods.
\newblock \url{https://github.com/huggingface/peft}.

\bibitem[{OpenAI(2023)}]{openai2023gpt4}
OpenAI. 2023.
\newblock \href {http://arxiv.org/abs/2303.08774} {Gpt-4 technical report}.

\bibitem[{Ott et~al.(2019)Ott, Edunov, Baevski, Fan, Gross, Ng, Grangier, and
  Auli}]{ott-etal-2019-fairseq}
Myle Ott, Sergey Edunov, Alexei Baevski, Angela Fan, Sam Gross, Nathan Ng,
  David Grangier, and Michael Auli. 2019.
\newblock \href {https://doi.org/10.18653/v1/N19-4009} {fairseq: A fast,
  extensible toolkit for sequence modeling}.
\newblock In \emph{Proceedings of the 2019 Conference of the North {A}merican
  Chapter of the Association for Computational Linguistics (Demonstrations)},
  pages 48--53, Minneapolis, Minnesota. Association for Computational
  Linguistics.

\bibitem[{Ouyang et~al.(2022{\natexlab{a}})Ouyang, Wu, Jiang, Almeida,
  Wainwright, Mishkin, Zhang, Agarwal, Slama, Ray, Schulman, Hilton, Kelton,
  Miller, Simens, Askell, Welinder, Christiano, Leike, and
  Lowe}]{LongOuyang2022TrainingLM}
Long Ouyang, Jeff Wu, Xu~Jiang, Diogo Almeida, Carroll Wainwright, Pamela
  Mishkin, Chong Zhang, Sandhini Agarwal, Katarina Slama, Alex Ray, John
  Schulman, Jacob Hilton, Fraser Kelton, Luke Miller, Maddie Simens, Amanda
  Askell, Peter Welinder, Paul Christiano, Jan Leike, and Ryan Lowe.
  2022{\natexlab{a}}.
\newblock Training language models to follow instructions with human feedback.

\bibitem[{Ouyang et~al.(2022{\natexlab{b}})Ouyang, Wu, Jiang, Almeida,
  Wainwright, Mishkin, Zhang, Agarwal, Slama, Gray, Schulman, Hilton, Kelton,
  Miller, Simens, Askell, Welinder, Christiano, Leike, and
  Lowe}]{ouyang2022training}
Long Ouyang, Jeffrey Wu, Xu~Jiang, Diogo Almeida, Carroll Wainwright, Pamela
  Mishkin, Chong Zhang, Sandhini Agarwal, Katarina Slama, Alex Gray, John
  Schulman, Jacob Hilton, Fraser Kelton, Luke Miller, Maddie Simens, Amanda
  Askell, Peter Welinder, Paul Christiano, Jan Leike, and Ryan Lowe.
  2022{\natexlab{b}}.
\newblock \href {https://openreview.net/forum?id=TG8KACxEON} {Training language
  models to follow instructions with human feedback}.
\newblock In \emph{Advances in Neural Information Processing Systems}.

\bibitem[{Radford et~al.(2019)Radford, Wu, Child, Luan, Amodei, and
  Sutskever}]{Radford2019LanguageMA}
Alec Radford, Jeff Wu, Rewon Child, David Luan, Dario Amodei, and Ilya
  Sutskever. 2019.
\newblock Language models are unsupervised multitask learners.

\bibitem[{Ren et~al.(2023)Ren, Zhang, Tian, and Li}]{ren2023crest}
Zhichu Ren, Zhen Zhang, Yunsheng Tian, and Ju~Li. 2023.
\newblock Crest--copilot for real-world experimental scientist.

\bibitem[{Scao et~al.(2022)Scao, Fan, Akiki, Pavlick, Ili{\'c}, Hesslow,
  Castagn{\'e}, Luccioni, Yvon, Gall{\'e} et~al.}]{scao2022bloom}
Teven~Le Scao, Angela Fan, Christopher Akiki, Ellie Pavlick, Suzana Ili{\'c},
  Daniel Hesslow, Roman Castagn{\'e}, Alexandra~Sasha Luccioni, Fran{\c{c}}ois
  Yvon, Matthias Gall{\'e}, et~al. 2022.
\newblock Bloom: A 176b-parameter open-access multilingual language model.
\newblock \emph{arXiv preprint arXiv:2211.05100}.

\bibitem[{Shazeer(2019)}]{shazeer2019fast}
Noam Shazeer. 2019.
\newblock \href {http://arxiv.org/abs/1911.02150} {Fast transformer decoding:
  One write-head is all you need}.

\bibitem[{Shoeybi et~al.(2020)Shoeybi, Patwary, Puri, LeGresley, Casper, and
  Catanzaro}]{shoeybi2020megatronlm}
Mohammad Shoeybi, Mostofa Patwary, Raul Puri, Patrick LeGresley, Jared Casper,
  and Bryan Catanzaro. 2020.
\newblock \href {http://arxiv.org/abs/1909.08053} {Megatron-lm: Training
  multi-billion parameter language models using model parallelism}.

\bibitem[{Su et~al.(2022)Su, Lu, Pan, Murtadha, Wen, and Liu}]{su2022roformer}
Jianlin Su, Yu~Lu, Shengfeng Pan, Ahmed Murtadha, Bo~Wen, and Yunfeng Liu.
  2022.
\newblock \href {http://arxiv.org/abs/2104.09864} {Roformer: Enhanced
  transformer with rotary position embedding}.

\bibitem[{Sun et~al.(2022)Sun, Dong, Patra, Ma, Huang, Benhaim, Chaudhary,
  Song, and Wei}]{sun2022lengthextrapolatable}
Yutao Sun, Li~Dong, Barun Patra, Shuming Ma, Shaohan Huang, Alon Benhaim,
  Vishrav Chaudhary, Xia Song, and Furu Wei. 2022.
\newblock \href {http://arxiv.org/abs/2212.10554} {A length-extrapolatable
  transformer}.

\bibitem[{Taori et~al.(2023)Taori, Gulrajani, Zhang, Dubois, Li, Guestrin,
  Liang, and Hashimoto}]{alpaca}
Rohan Taori, Ishaan Gulrajani, Tianyi Zhang, Yann Dubois, Xuechen Li, Carlos
  Guestrin, Percy Liang, and Tatsunori~B. Hashimoto. 2023.
\newblock Stanford alpaca: An instruction-following llama model.
\newblock \url{https://github.com/tatsu-lab/stanford_alpaca}.

\bibitem[{Touvron et~al.(2023{\natexlab{a}})Touvron, Lavril, Izacard, Martinet,
  Lachaux, Lacroix, Rozière, Goyal, Hambro, Azhar, Rodriguez, Joulin, Grave,
  and Lample}]{touvron2023llama}
Hugo Touvron, Thibaut Lavril, Gautier Izacard, Xavier Martinet, Marie-Anne
  Lachaux, Timothée Lacroix, Baptiste Rozière, Naman Goyal, Eric Hambro,
  Faisal Azhar, Aurelien Rodriguez, Armand Joulin, Edouard Grave, and Guillaume
  Lample. 2023{\natexlab{a}}.
\newblock \href {http://arxiv.org/abs/2302.13971} {Llama: Open and efficient
  foundation language models}.

\bibitem[{Touvron et~al.(2023{\natexlab{b}})Touvron, Martin, Stone, Albert,
  Almahairi, Babaei, Bashlykov, Batra, Bhargava, Bhosale, Bikel, Blecher,
  Ferrer, Chen, Cucurull, Esiobu, Fernandes, Fu, Fu, Fuller, Gao, Goswami,
  Goyal, Hartshorn, Hosseini, Hou, Inan, Kardas, Kerkez, Khabsa, Kloumann,
  Korenev, Koura, Lachaux, Lavril, Lee, Liskovich, Lu, Mao, Martinet, Mihaylov,
  Mishra, Molybog, Nie, Poulton, Reizenstein, Rungta, Saladi, Schelten, Silva,
  Smith, Subramanian, Tan, Tang, Taylor, Williams, Kuan, Xu, Yan, Zarov, Zhang,
  Fan, Kambadur, Narang, Rodriguez, Stojnic, Edunov, and
  Scialom}]{touvron2023llama2}
Hugo Touvron, Louis Martin, Kevin Stone, Peter Albert, Amjad Almahairi, Yasmine
  Babaei, Nikolay Bashlykov, Soumya Batra, Prajjwal Bhargava, Shruti Bhosale,
  Dan Bikel, Lukas Blecher, Cristian~Canton Ferrer, Moya Chen, Guillem
  Cucurull, David Esiobu, Jude Fernandes, Jeremy Fu, Wenyin Fu, Brian Fuller,
  Cynthia Gao, Vedanuj Goswami, Naman Goyal, Anthony Hartshorn, Saghar
  Hosseini, Rui Hou, Hakan Inan, Marcin Kardas, Viktor Kerkez, Madian Khabsa,
  Isabel Kloumann, Artem Korenev, Punit~Singh Koura, Marie-Anne Lachaux,
  Thibaut Lavril, Jenya Lee, Diana Liskovich, Yinghai Lu, Yuning Mao, Xavier
  Martinet, Todor Mihaylov, Pushkar Mishra, Igor Molybog, Yixin Nie, Andrew
  Poulton, Jeremy Reizenstein, Rashi Rungta, Kalyan Saladi, Alan Schelten, Ruan
  Silva, Eric~Michael Smith, Ranjan Subramanian, Xiaoqing~Ellen Tan, Binh Tang,
  Ross Taylor, Adina Williams, Jian~Xiang Kuan, Puxin Xu, Zheng Yan, Iliyan
  Zarov, Yuchen Zhang, Angela Fan, Melanie Kambadur, Sharan Narang, Aurelien
  Rodriguez, Robert Stojnic, Sergey Edunov, and Thomas Scialom.
  2023{\natexlab{b}}.
\newblock \href {http://arxiv.org/abs/2307.09288} {Llama 2: Open foundation and
  fine-tuned chat models}.

\bibitem[{Tworkowski et~al.(2023)Tworkowski, Staniszewski, Pacek, Wu,
  Michalewski, and Miłoś}]{tworkowski2023focused}
Szymon Tworkowski, Konrad Staniszewski, Mikołaj Pacek, Yuhuai Wu, Henryk
  Michalewski, and Piotr Miłoś. 2023.
\newblock \href {http://arxiv.org/abs/2307.03170} {Focused transformer:
  Contrastive training for context scaling}.

\bibitem[{Vaswani et~al.(2017)Vaswani, Shazeer, Parmar, Uszkoreit, Jones,
  Gomez, Kaiser, and Polosukhin}]{vaswani2017attention}
Ashish Vaswani, Noam Shazeer, Niki Parmar, Jakob Uszkoreit, Llion Jones,
  Aidan~N. Gomez, Lukasz Kaiser, and Illia Polosukhin. 2017.
\newblock \href {http://arxiv.org/abs/1706.03762} {Attention is all you need}.

\bibitem[{Wang et~al.(2023{\natexlab{a}})Wang, Qin, Jacobs, Holmes,
  Rajbhandari, Ruwase, Yan, Yang, and He}]{wang2023zero}
Guanhua Wang, Heyang Qin, Sam~Ade Jacobs, Connor Holmes, Samyam Rajbhandari,
  Olatunji Ruwase, Feng Yan, Lei Yang, and Yuxiong He. 2023{\natexlab{a}}.
\newblock \href {http://arxiv.org/abs/2306.10209} {Zero++: Extremely efficient
  collective communication for giant model training}.

\bibitem[{Wang et~al.(2023{\natexlab{b}})Wang, Xie, Jiang, Mandlekar, Xiao,
  Zhu, Fan, and Anandkumar}]{wang2023voyager}
Guanzhi Wang, Yuqi Xie, Yunfan Jiang, Ajay Mandlekar, Chaowei Xiao, Yuke Zhu,
  Linxi Fan, and Anima Anandkumar. 2023{\natexlab{b}}.
\newblock \href {http://arxiv.org/abs/2305.16291} {Voyager: An open-ended
  embodied agent with large language models}.

\bibitem[{Wang et~al.(2023{\natexlab{c}})Wang, Zhang, Yang, Shi, Zhou, Hao,
  Xiong, Li, Sim, Chen, Zhu, Yang, Nik, Liu, Lin, Wang, Liu, Chen, Xu, Liu,
  Guo, and Fu}]{wang2023interactive}
Zekun Wang, Ge~Zhang, Kexin Yang, Ning Shi, Wangchunshu Zhou, Shaochun Hao,
  Guangzheng Xiong, Yizhi Li, Mong~Yuan Sim, Xiuying Chen, Qingqing Zhu,
  Zhenzhu Yang, Adam Nik, Qi~Liu, Chenghua Lin, Shi Wang, Ruibo Liu, Wenhu
  Chen, Ke~Xu, Dayiheng Liu, Yike Guo, and Jie Fu. 2023{\natexlab{c}}.
\newblock \href {http://arxiv.org/abs/2305.13246} {Interactive natural language
  processing}.

\bibitem[{Wei et~al.(2021)Wei, Bosma, Zhao, Guu, Yu, Lester, Du, Dai, and
  Le}]{weifinetuned}
Jason Wei, Maarten Bosma, Vincent Zhao, Kelvin Guu, Adams~Wei Yu, Brian Lester,
  Nan Du, Andrew~M Dai, and Quoc~V Le. 2021.
\newblock Finetuned language models are zero-shot learners.
\newblock In \emph{International Conference on Learning Representations}.

\bibitem[{Wei et~al.(2022{\natexlab{a}})Wei, Tay, Bommasani, Raffel, Zoph,
  Borgeaud, Yogatama, Bosma, Zhou, Metzler et~al.}]{wei2022emergent}
Jason Wei, Yi~Tay, Rishi Bommasani, Colin Raffel, Barret Zoph, Sebastian
  Borgeaud, Dani Yogatama, Maarten Bosma, Denny Zhou, Donald Metzler, et~al.
  2022{\natexlab{a}}.
\newblock Emergent abilities of large language models.
\newblock \emph{arXiv preprint arXiv:2206.07682}.

\bibitem[{Wei et~al.(2022{\natexlab{b}})Wei, Wang, Schuurmans, Bosma, Chi, Le,
  and Zhou}]{wei2022chain}
Jason Wei, Xuezhi Wang, Dale Schuurmans, Maarten Bosma, Ed~Chi, Quoc Le, and
  Denny Zhou. 2022{\natexlab{b}}.
\newblock Chain of thought prompting elicits reasoning in large language
  models.
\newblock \emph{arXiv preprint arXiv:2201.11903}.

\bibitem[{Weng et~al.(2023{\natexlab{a}})Weng, Li, Xia, Zhu, Sun, He, Liu, and
  Zhao}]{Weng2023LargeLM}
Yixuan Weng, Bin Li, Fei Xia, Minjun Zhu, Bing Sun, Shizhu He, Kang Liu, and
  Jun Zhao. 2023{\natexlab{a}}.
\newblock Large language models need holistically thought in medical
  conversational qa.
\newblock \emph{ArXiv}, abs/2305.05410.

\bibitem[{Weng et~al.(2022)Weng, Zhu, He, Liu, and Zhao}]{weng2022large}
Yixuan Weng, Minjun Zhu, Shizhu He, Kang Liu, and Jun Zhao. 2022.
\newblock Large language models are reasoners with self-verification.
\newblock \emph{arXiv preprint arXiv:2212.09561}.

\bibitem[{Weng et~al.(2023{\natexlab{b}})Weng, Zhu, Xia, Li, He, Liu, and
  Zhao}]{weng2023neural}
Yixuan Weng, Minjun Zhu, Fei Xia, Bin Li, Shizhu He, Kang Liu, and Jun Zhao.
  2023{\natexlab{b}}.
\newblock Neural comprehension: Language models with compiled neural networks.
\newblock \emph{arXiv preprint arXiv:2304.01665}.

\bibitem[{Wolf et~al.(2020)Wolf, Debut, Sanh, Chaumond, Delangue, Moi, Cistac,
  Rault, Louf, Funtowicz, Davison, Shleifer, von Platen, Ma, Jernite, Plu, Xu,
  Le~Scao, Gugger, Drame, Lhoest, and Rush}]{wolf-etal-2020-transformers}
Thomas Wolf, Lysandre Debut, Victor Sanh, Julien Chaumond, Clement Delangue,
  Anthony Moi, Pierric Cistac, Tim Rault, Remi Louf, Morgan Funtowicz, Joe
  Davison, Sam Shleifer, Patrick von Platen, Clara Ma, Yacine Jernite, Julien
  Plu, Canwen Xu, Teven Le~Scao, Sylvain Gugger, Mariama Drame, Quentin Lhoest,
  and Alexander Rush. 2020.
\newblock \href {https://doi.org/10.18653/v1/2020.emnlp-demos.6} {Transformers:
  State-of-the-art natural language processing}.
\newblock In \emph{Proceedings of the 2020 Conference on Empirical Methods in
  Natural Language Processing: System Demonstrations}, pages 38--45, Online.
  Association for Computational Linguistics.

\bibitem[{Xi et~al.(2023)Xi, Li, Chen, and Zhu}]{xi2023training}
Haocheng Xi, Changhao Li, Jianfei Chen, and Jun Zhu. 2023.
\newblock \href {http://arxiv.org/abs/2306.11987} {Training transformers with
  4-bit integers}.

\bibitem[{Xia et~al.(2022{\natexlab{a}})Xia, Li, Weng, He, Liu, Sun, Li, and
  Zhao}]{xia-etal-2022-medconqa}
Fei Xia, Bin Li, Yixuan Weng, Shizhu He, Kang Liu, Bin Sun, Shutao Li, and Jun
  Zhao. 2022{\natexlab{a}}.
\newblock \href {https://aclanthology.org/2022.emnlp-demos.15} {{M}ed{C}on{QA}:
  Medical conversational question answering system based on knowledge graphs}.
\newblock In \emph{Proceedings of the The 2022 Conference on Empirical Methods
  in Natural Language Processing: System Demonstrations}, pages 148--158, Abu
  Dhabi, UAE. Association for Computational Linguistics.

\bibitem[{Xia et~al.(2022{\natexlab{b}})Xia, Li, Weng, He, Liu, Sun, Li, and
  Zhao}]{xia2022medconqa}
Fei Xia, Bin Li, Yixuan Weng, Shizhu He, Kang Liu, Bin Sun, Shutao Li, and Jun
  Zhao. 2022{\natexlab{b}}.
\newblock Medconqa: Medical conversational question answering system based on
  knowledge graphs.
\newblock In \emph{Proceedings of the The 2022 Conference on Empirical Methods
  in Natural Language Processing: System Demonstrations}, pages 148--158.

\bibitem[{Xu et~al.(2023)Xu, Guo, Duan, and McAuley}]{xu2023baize}
Canwen Xu, Daya Guo, Nan Duan, and Julian McAuley. 2023.
\newblock Baize: An open-source chat model with parameter-efficient tuning on
  self-chat data.
\newblock \emph{arXiv preprint arXiv:2304.01196}.

\bibitem[{Yao et~al.(2023)Yao, Zhao, Yu, Du, Shafran, Narasimhan, and
  Cao}]{yao2023react}
Shunyu Yao, Jeffrey Zhao, Dian Yu, Nan Du, Izhak Shafran, Karthik Narasimhan,
  and Yuan Cao. 2023.
\newblock \href {http://arxiv.org/abs/2210.03629} {React: Synergizing reasoning
  and acting in language models}.

\bibitem[{Zeng et~al.(2022)Zeng, Liu, Du, Wang, Lai, Ding, Yang, Xu, Zheng, Xia
  et~al.}]{zeng2022glm}
Aohan Zeng, Xiao Liu, Zhengxiao Du, Zihan Wang, Hanyu Lai, Ming Ding, Zhuoyi
  Yang, Yifan Xu, Wendi Zheng, Xiao Xia, et~al. 2022.
\newblock Glm-130b: An open bilingual pre-trained model.
\newblock \emph{arXiv preprint arXiv:2210.02414}.

\bibitem[{Zeng et~al.(2020)Zeng, Yang, Ju, Yang, Wang, Zhang, Zhou, Zeng, Dong,
  Zhang et~al.}]{zeng2020meddialog}
Guangtao Zeng, Wenmian Yang, Zeqian Ju, Yue Yang, Sicheng Wang, Ruisi Zhang,
  Meng Zhou, Jiaqi Zeng, Xiangyu Dong, Ruoyu Zhang, et~al. 2020.
\newblock Meddialog: Large-scale medical dialogue dataset.
\newblock In \emph{Proceedings of the 2020 Conference on Empirical Methods in
  Natural Language Processing (EMNLP)}.

\bibitem[{Zhang et~al.(2022)Zhang, Roller, Goyal, Artetxe, Chen, Chen, Dewan,
  Diab, Li, Lin et~al.}]{zhang2022opt}
Susan Zhang, Stephen Roller, Naman Goyal, Mikel Artetxe, Moya Chen, Shuohui
  Chen, Christopher Dewan, Mona Diab, Xian Li, Xi~Victoria Lin, et~al. 2022.
\newblock Opt: Open pre-trained transformer language models.
\newblock \emph{arXiv preprint arXiv:2205.01068}.

\bibitem[{Zhao et~al.(2023{\natexlab{a}})Zhao, Zhou, Li, Tang, Wang, Hou, Min,
  Zhang, Zhang, Dong, Du, Yang, Chen, Chen, Jiang, Ren, Li, Tang, Liu, Liu,
  Nie, and Wen}]{zhao2023survey}
Wayne~Xin Zhao, Kun Zhou, Junyi Li, Tianyi Tang, Xiaolei Wang, Yupeng Hou,
  Yingqian Min, Beichen Zhang, Junjie Zhang, Zican Dong, Yifan Du, Chen Yang,
  Yushuo Chen, Zhipeng Chen, Jinhao Jiang, Ruiyang Ren, Yifan Li, Xinyu Tang,
  Zikang Liu, Peiyu Liu, Jian-Yun Nie, and Ji-Rong Wen. 2023{\natexlab{a}}.
\newblock \href {http://arxiv.org/abs/2303.18223} {A survey of large language
  models}.

\bibitem[{Zhao et~al.(2023{\natexlab{b}})Zhao, Li, Hou, Zhao, Tian, Liu, Chen,
  Sun, Liu, Mao, Guo, Gou, Wu, Zhu, Shi, Chen, Huang, Chen, Liu, Li, Chen, Sun,
  Kang, Du, Shen, and Yan}]{zhao-etal-2023-tencentpretrain}
Zhe Zhao, Yudong Li, Cheng Hou, Jing Zhao, Rong Tian, Weijie Liu, Yiren Chen,
  Ningyuan Sun, Haoyan Liu, Weiquan Mao, Han Guo, Weigang Gou, Taiqiang Wu, Tao
  Zhu, Wenhang Shi, Chen Chen, Shan Huang, Sihong Chen, Liqun Liu, Feifei Li,
  Xiaoshuai Chen, Xingwu Sun, Zhanhui Kang, Xiaoyong Du, Linlin Shen, and Kimmo
  Yan. 2023{\natexlab{b}}.
\newblock \href {https://aclanthology.org/2023.acl-demo.20}
  {{T}encent{P}retrain: A scalable and flexible toolkit for pre-training models
  of different modalities}.
\newblock In \emph{Proceedings of the 61st Annual Meeting of the Association
  for Computational Linguistics (Volume 3: System Demonstrations)}, pages
  217--225, Toronto, Canada. Association for Computational Linguistics.

\bibitem[{Zhou et~al.(2023)Zhou, Liu, Xu, Iyer, Sun, Mao, Ma, Efrat, Yu, Yu,
  Zhang, Ghosh, Lewis, Zettlemoyer, and Levy}]{zhou2023lima}
Chunting Zhou, Pengfei Liu, Puxin Xu, Srini Iyer, Jiao Sun, Yuning Mao, Xuezhe
  Ma, Avia Efrat, Ping Yu, Lili Yu, Susan Zhang, Gargi Ghosh, Mike Lewis, Luke
  Zettlemoyer, and Omer Levy. 2023.
\newblock \href {http://arxiv.org/abs/2305.11206} {Lima: Less is more for
  alignment}.

\bibitem[{Zhu et~al.(2023)Zhu, Weng, He, Liu, and Zhao}]{zhu2023towards}
Minjun Zhu, Yixuan Weng, Shizhu He, Kang Liu, and Jun Zhao. 2023.
\newblock Towards graph-hop retrieval and reasoning in complex question
  answering over textual database.
\newblock \emph{arXiv preprint arXiv:2305.14211}.

\end{thebibliography}
\bibliographystyle{acl_natbib}

\appendix

\section{Support Models}
\label{appendix:models}

LMTuner enables rapid deployment of large language models ranging from 300 million to 130 billion parameters, facilitating their application to textual data tasks. We list the models supported for training in LMTuner in Table \ref{table:models}. (Please note that some of these models are not fully allowed for commercial use.)
\begin{table*}[t]
\resizebox{\textwidth}{!}{%
\centering
\begin{tabular}{c|c|c|c|c|c}
\toprule
Model Size &  \ \ \ \ 16-bit Finetune \ \ \ \ & \ \ \ \ 16-bit LOMO\ \ \ \ & \ \ \ \ 16-bit LoRA\ \ \ \  &\ \ \ \ 8-bit LoRA\ \ \ \ &\ \ \ \ 4-bit LoRA\ \ \ \ \\ \hline
<= 1B & 8 GB& 4 GB&4 GB&6 GB&6 GB\\
7B & 8 GB & 6 GB & 6 GB & 8 GB & 8 GB \\
13B & 16 GB & 10 GB & 10 GB & 12 GB & 10 GB \\
33B & 2x48 GB & 24 GB & 24 GB & 32 GB & 32 GB \\
70B & 8x24 GB\quad4x48 GB\quad 2x80 GB & 2x24 GB\quad 48 GB & 2x24 GB\quad 48 GB & 2x32 GB\quad 80 GB & 2x24 GB\quad 48 GB \\
130B & 8x48 GB\quad4x80 GB & 8x24 GB\quad4x48 GB\quad2x80 GB & 8x24 GB\quad4x48 GB\quad2x80 GB & 8x24 GB\quad4x48 GB\quad2x80 GB & 8x24 GB\quad4x48 GB\quad2x80 GB \\
\bottomrule
\end{tabular}}
\caption{The minimum configuration required for different model size in LMTuner (when Batchsize=1). We suggest appropriately increasing the number of GPUs to accelerate training the models.}
\label{table:models}
\end{table*}

\begin{table}[h]
\resizebox{0.45\textwidth}{!}{%
\centering
\begin{tabular}{lr}
\toprule
GPT-2 & \cite{Radford2019LanguageMA}\\
GPT-Neo-1.3B & \cite{gpt-neo}\\
ChatGLM-6B & \cite{du2022glm}\\
ChatGLM2-6B & \cite{du2022glm}\\
Llama-7B & \cite{touvron2023llama}\\
Llama-13B & \cite{touvron2023llama}\\
Llama-33B & \cite{touvron2023llama}\\
Llama-65B & \cite{touvron2023llama}\\
Llama2-7B & \cite{touvron2023llama2}\\
Llama2-13B & \cite{touvron2023llama2}\\
Llama2-70B & \cite{touvron2023llama2}\\
GLM-130B & \cite{zeng2022glm}\\
\bottomrule
\end{tabular}}
\caption{LMTuner supports pre-trained language models for rapid deployment.}
\label{table:models}
\end{table}

LMTuner also provides capabilities to load pretrained models from Transformers, however this necessitates redefining the data tokenizer. In addition, LMTuner facilitates training models from scratch to incorporate specialized architectures such as FlashAttention and Multi-Query Attention through its flexible hook. The modular hook-based approach streamlines implementing customized attentional mechanisms and training objectives. This balancing of usability and flexibility makes LMTuner a promising platform for rapidly prototyping.

\section{Support Datasets}
We have supported multiple QA datasets in different fields for quick training. The provision of ready-to-use domain-specific datasets eliminates the need for users to invest significant time and effort in dataset curation and preprocessing. With high-quality training data covering key domains, users can promptly initialize model training and optimization workflows. 

\begin{itemize}
    \item For English datasets, we default to loading the LIMA\footnote{\url{https://huggingface.co/datasets/GAIR/lima}} \cite{zhou2023lima} dataset, which includes 1,030 diverse and high-quality QA samples.
    \item For Chinese datasets, we manually translated the LIMA dataset and added an additional 60 Chinese samples covering different fields such as Chinese history, Marxism, essay writing, and Chinese humor, making it a QA dataset with 1,090 samples covering multiple fields and Chinese characteristics.
    \item For Chinese Medical datasets, We selected about 60,000 medical consultation dialogues from CMCQA \cite{xia2022medconqa}, which contained QA samples from over 30 different departments, including pediatrics, gynecology, internal medicine, oncology and more.
\end{itemize}

\section{LMTuner's GPU Memory Usage}
Normally, training LLMs requires a huge amount of GPU memory. Recent advancements in EFT techniques such as LoRA and LOMO have enabled the training of extremely large language models on consumer GPUs with as little as 24GB of memory. In addition, tensor parallelism techniques can split a model across different GPUs, making it possible to scale up training of even larger LLMs. In Table 3 we list the minimum recommended GPU memory configuration by LMTuner for training models of different sizes.

\section{System Configuration and Requirements}

LMTuner (\url{https://github.com/WENGSYX/LMTuner}) is an open-source system that offers a command-line interface for training LLMs,  without requiring any coding experience. The system requirements
 include: Ubuntu 14.04+, Debian 8+, CentOS 6+, or Fedora 27+.  And an NVIDIA GPU with driver version >= 460.32.03 or AMD GPU with ROMc >= 4.0.

 In addition, a series of Python libraries, including Apex\footnote{\url{https://github.com/NVIDIA/apex}}, Pytorch\footnote{\url{https://pytorch.org/}} and DeepSpeed \footnote{\url{https://github.com/microsoft/DeepSpeed}}, are also required. For details, please refer to \url{https://wengsyx.github.io/LMTuner/install.html}. 

 After installation is complete, you can directly launch via Command-Line Interface without modifying any code by using the one-line code \raisebox{-0.25\height}{\includegraphics[width=50pt]{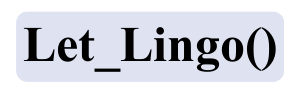}}.

LMTuner requires the provision of an OpenAI Key to facilitate the collaborative determination of training parameters with GPT-4 through dialogue. However, we have also pre-defined ten sets of configuration questions related to training LLMs in a format similar to a questionnaire, to enable users in countries that do not support GPT-4 to quickly initiate LMTuner \footnote{Take a look at the current list of Supported Countries and Territories for OpenAI: \url{https://platform.openai.com/docs/supported-countries}}.

Finally, LMTuner will automatically generate and run a command line to invoke deepspeed according to the requirements without needing to modify at the code level. This not only simplifies the training process, but also facilitates modifying and recording the configuration of hyperparameters, helping users without basic knowledge to also train LLMs.

\end{document}